\documentclass[sigconf]{acmart}
\AtBeginDocument{%
  }

\copyrightyear{2025}
\acmYear{2025}
\setcopyright{rightsretained}
\acmConference[ICMR '25]{Proceedings of the 2025 International Conference on Multimedia Retrieval}{June 30-July 3, 2025}{Chicago, IL, USA}
\acmBooktitle{Proceedings of the 2025 International Conference on Multimedia Retrieval (ICMR '25), June 30-July 3, 2025, Chicago, IL, USA}
\acmDOI{10.1145/3731715.3733310}
\acmISBN{979-8-4007-1877-9/2025/06}

\usepackage{multirow}

\graphicspath{{./fig/}}

\newcommand{\myPara}[1]{\noindent\textbf{#1}}

\newcommand{\bl}[1]{\textbf{#1}}

\newcommand{\mc}[1]{\mathcal{#1}}

\begin{document}

\title{Divide and Conquer: Static-Dynamic Collaboration for Few-Shot Class-Incremental Learning}





\author{Kexin Bao}
\orcid{0000-0003-4921-6112}
\affiliation{%
 \institution{Institute of Information Engineering, Chinese Academy of Sciences}
  \institution{School of Cyber Security, University of Chinese Academy of Sciences}
  \city{Beijing}
  \country{China}}
\email{baokexin@iie.ac.cn}

\author{Daichi Zhang}
\orcid{0000-0002-5377-964X}
\authornote{Shiming Ge and Daichi Zhang are corresponding authors.}
\affiliation{%
  \institution{Institute of Information Engineering, Chinese Academy of Sciences}
  \institution{School of Cyber Security, University of Chinese Academy of Sciences}
  \city{Beijing}
  \country{China}}
\email{zhangdaichi@iie.ac.cn}

\author{Yong Li}
\affiliation{%
 \institution{Institute of Information Engineering, Chinese Academy of Sciences}
  \city{Beijing}
  \country{China}}
\email{liyong@iie.ac.cn} 

\author{Dan Zeng}
\orcid{0000-0003-1300-1769}
\affiliation{%
 \institution{Department of Communication Engineering, Shanghai University}
  \city{Shanghai}
  \country{China}}
\email{dzeng@shu.edu.cn} 

\author{Shiming Ge}
\orcid{0000-0001-5293-310X}
\authornotemark[1]
\affiliation{%
 \institution{Institute of Information Engineering, Chinese Academy of Sciences}
  \city{Beijing}
  \country{China}}
\email{geshiming@iie.ac.cn} 







\renewcommand{\shortauthors}{Kexin Bao, Daichi Zhang, Yong Li, Dan Zeng and Shiming Ge.}

\begin{abstract}

Few-shot class-incremental learning (FSCIL) aims to continuously recognize novel classes under limited data, which suffers from the key stability-plasticity dilemma: balancing the retention of old knowledge with the acquisition of new knowledge. To address this issue, we divide the task into two different stages and propose a framework termed Static-Dynamic Collaboration (\textbf{SDC}) to achieve a better trade-off between stability and plasticity. Specifically, our method divides the normal pipeline of FSCIL into Static Retaining Stage (\textbf{SRS}) and Dynamic Learning Stage (\textbf{DLS}), which harnesses old static and incremental dynamic class information, respectively. During SRS, we train an initial model with sufficient data in the base session and preserve the key part as static memory to retain fundamental old knowledge. During DLS, we introduce an extra dynamic projector jointly trained with the previous static memory. By employing both stages, our method achieves improved retention of old knowledge while continuously adapting to new classes. Extensive experiments on three public benchmarks and a real-world application dataset demonstrate that our method achieves state-of-the-art performance against other competitors.

\end{abstract}




\maketitle

\section{Introduction}

\begin{figure}[t]
\centering
\includegraphics[width=1\linewidth]{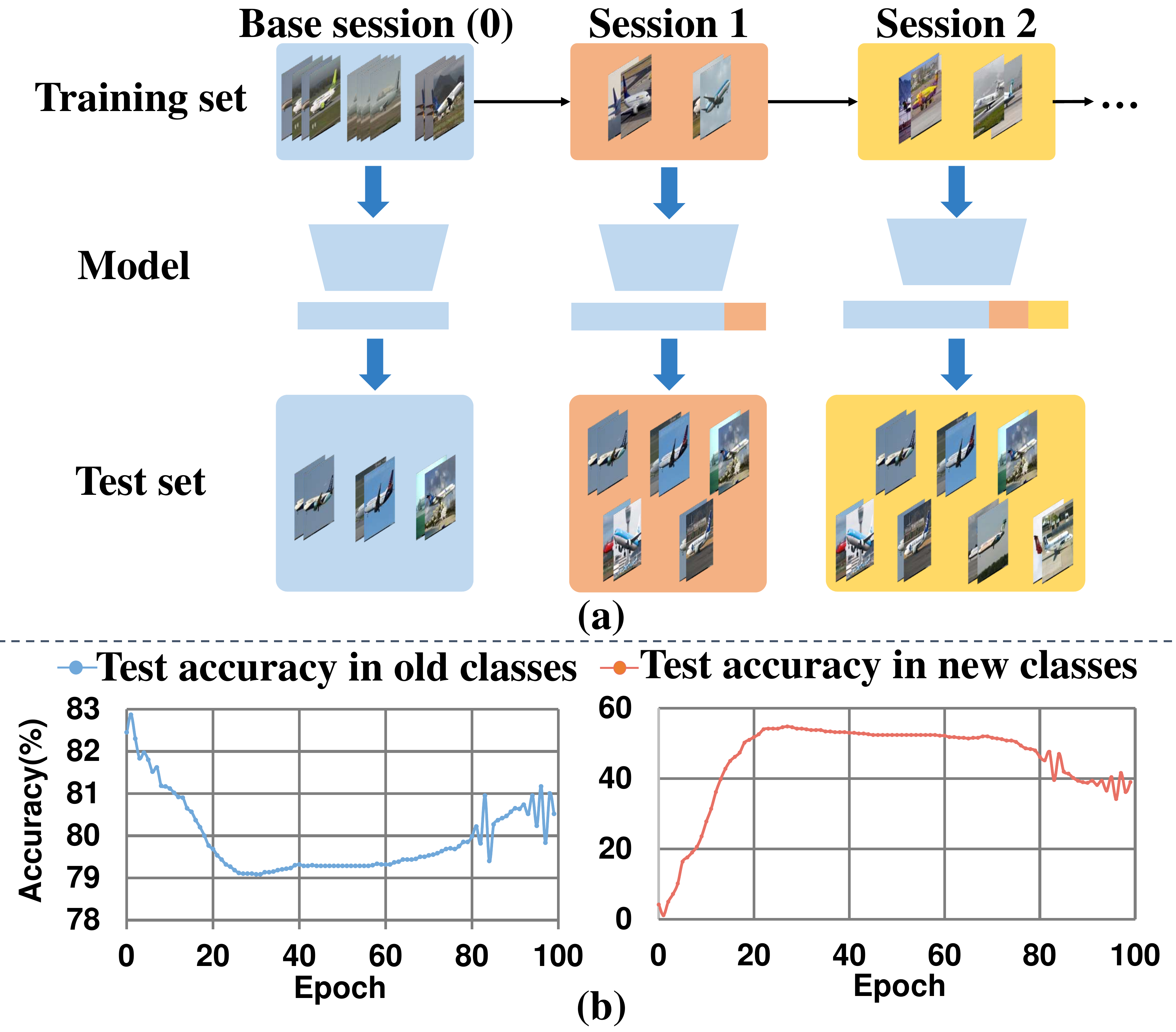}
\caption{(a) The setting of few-shot class-incremental learning, which aims to recognize new classes without forgetting old ones. (b) The phenomenon of imbalance between preserving old classes and adapting to new classes. }
\Description{Motivation}
\label{fig:motivation}
\end{figure}

Deep neural networks (DNNs) have obtained significant breakthroughs across various tasks with extensive annotated data, abundant computing resources and plenty of time~\cite{DBLP:conf/iccv/FuYGX023,DBLP:conf/aaai/0001SYZ24,DBLP:conf/aaai/0001WZ24}. However, in real-world scenarios, data tends to come in a continuous stream due to new classes emerging regularly, posing challenges for traditional DNNs that rely on static datasets. Class-incremental learning (CIL) enables a model to absorb new category knowledge phase-by-phase with the increased data~\cite{lange2021tpami}, which does not retrain the model whenever adding novel classes. Yet, collecting and labeling an adequate amount of data is time-consuming and labor-intensive, particularly in some data-limited contexts due to scarcity, expensive, or privacy concerns (e.g., automated surveillance systems~\cite{DBLP:journals/access/HassanDMM24,DBLP:journals/access/OsokinRMPBO24}, intelligent medicine~\cite{DBLP:journals/access/AlamSAACJ24}, autonomous driving~\cite{DBLP:journals/access/TaherinavidMCYKP24}, and automated surveillance~\cite{DBLP:journals/access/PietaJBJPLDJOWIL24}). And few-shot class-incremental learning (FSCIL) forces a model to learn novel concepts given a very restricted quantity of labeled samples facing a continuous stream~\cite{tao2020cvpr}, as illustrated in Figure~\ref{fig:motivation} (a). Similar to the CIL\cite{DBLP:conf/cvpr/KimH23}, FSCIL encounters the stability-plasticity dilemma: balancing the retention of old knowledge with the acquisition of new knowledge under data scarcity. This dilemma is more pronounced in FSCIL than in traditional CIL settings due to the limited data available for new classes.

Early works update the whole model using topology and knowledge distillation~\cite{tao2020cvpr,cheraghian2021cvpr,zhao2023cvpr}. However, updating a large number of variable parameters often dilutes or even discards previously learned knowledge, making it challenging to effectively resolve the dilemma. 
Mainstream methods freeze the backbone and finetune other parts of the model in incremental sessions~\cite{ji2023tip,chen2021iclr,rebuffi2017cvpr,lin2023tcsvt}. According to the different entry points, the mainstream methods can be divided into two groups. The first group makes efforts in the base session to stability allocate class space for incoming classes, including transferring the space of old classes~\cite{zou2022nips,kim2023iclr}, establishing virtual classes~\cite{zhou2022cvpr,song2023cvpr,lin2024aaai,zhou2022tpami}, and pre-allocating a fixed classifier~\cite{peng2022eccv,yang2023iclr}. 
The second group tends to extend networks~\cite{yang2023tpami,liu2023tpami,ji2023tip}, use an extra memory~\cite{ji2023tip,li2024tmm,lin2023tcsvt}, or constrain parameters via regularization~\cite{liu2023tpami,zhao2024tpami} in incremental sessions to adapt to new knowledge and maintain old knowledge flexibly. Although the first group emphasizes stability, the second group prioritizes plasticity, constraining the model with real samples. It possesses an advantage over the first group of methods in it can capture the authentic data distribution and characteristics more effectively, which is widely studied. However, existing methods usually suffer from the imbalance during incremental sessions, as shown in Fig.~\ref{fig:motivation} (b): the performance of old and new classes exhibits extreme imbalance during training. Thus, we raise the question: how to achieve a better trade-off between preserving old knowledge and adapting to new classes? 

To address that, we propose a framework termed Static-Dynamic Collaboration (\textbf{SDC}) to balance stability and plasticity, which consists of a \textbf{Static Retaining Stage (SRS)} and a \textbf{Dynamic Learning Stage (DLS)}. We build a model composed of a backbone and a classifier (including a projector and a fully connected (FC) layer). During SRS, we train the model with sufficient data in the base session, and store the projector as a static component. During DLS, we freeze the backbone and train a new dynamic projector in incremental sessions. The dynamic projector and FC layer are fine-tuned jointly with the static projector and an external memory. The static projector ensures stability by preserving foundational knowledge, while the dynamic projector adapts to new knowledge, promoting plasticity. This interplay allows SDC to maintain high continuity and discrimination, enabling the model to achieve high accuracy and adapt continuously to evolving data. Besides, we provide a brief analysis based on mutual information to demonstrate the interplay.

Our main contributions are threefold: 
\begin{itemize}
    \item We propose a framework termed Static-Dynamic Collaboration (\textbf{SDC}), to achieve high continuity and high discrimination. We briefly demonstrate our method can achieve a better trade-off between retaining old and adapting to new classes from the perspective of mutual information. 
    \item We introduce the Static Retaining Stage (\textbf{SRS}) and the Dynamic Learning Stage (\textbf{DLS}) to harness old static and incremental dynamic class information, ensuring that knowledge is preserved long-lasting and adaptable to changing contexts.
    \item Extensive experiments on three public benchmark datasets and a real-world application dataset demonstrate that our method outperforms other state-of-the-art competitors, supporting its stability and plasticity.
\end{itemize}


\section{Related Works}

\subsection{Class-Incremental Learning}
Class-incremental learning (CIL)trains a model for fitting a continuous feed of new emerging data~\cite{rebuffi2017cvpr,strubell2019acl,DBLP:journals/corr/abs-2004-08900}, suffering from catastrophic forgetting. Some works penalize parameters to minimize the impact of weights while training on new tasks~\cite{lee2020cvpr,Kirkpatrick_2017,DBLP:conf/eccv/AljundiBERT18}, mitigating catastrophic forgetting. Some methods invite knowledge distillation to compensate for the previous information in continuous tasks~\cite{10.5555/3454287.3455422,lin2023supervised,DBLP:conf/eccv/GaoZGZ22}. Furthermore, dynamic network methods extend the model to preserve knowledge, such as duplicating the model~\cite{DBLP:conf/cvpr/AljundiCT17}, fixing parameters~\cite{schwarz2018icml}, and fusing a dynamic architecture with exemplars~\cite{ebrahimi2020eccv}. Some methods replay data to reduce catastrophic forgetting~\cite{rebuffi2017cvpr,alijundi2019nips}. Given memory limitation and privacy issues, some methods exploit the pseudo-feature~\cite{DBLP:conf/iclr/ChaudhryRRE19,zhu2021cvpr}, which repeats the consolidation of past knowledge to strengthen the memory of what one has already learned. Most class incremental learning methods focus on incremental tasks and work to mitigate catastrophic forgetting through constraints, architecture, or pseudo-features, where the model has enough samples to learn new knowledge.

\subsection{Few-Shot Class-Incremental Learning}

Few-shot class-incremental learning (FSCIL) is proposed to learn with a few examples of new classes incrementally~\cite{tao2020cvpr}. Some early methods update the whole model using topology or knowledge distillation~\cite{tao2020cvpr,cheraghian2021cvpr}, which always suffers from catastrophic forgetting. To handle these challenges, mainstream methods freeze the backbone and fine-tune part of the model in incremental sessions, which we divide into two aspects: 1) pre-allocate spaces in the base session, and 2) constrain the model in incremental sessions.

\myPara{Pre-allocate spaces in FSCIL.} This group of methods preserves feature spaces for novel classes in the base session, which consists of using visual classes and widening the class space. Some methods use virtual classes to preserve space for new classes in advance. FACT~\cite{zhou2022cvpr} introduces virtual prototypes, MICS~\cite{DBLP:conf/wacv/KimJPY24} uses soft labels for virtual samples, and LIMIT~\cite{zhou2022tpami} synthesizes fake FSCIL tasks from the base dataset. However, although virtual classes reserve more learning spaces for incoming new classes, the model may encounter confusion due to the difference between the feature distribution of the new classes and the virtual classes. And some methods widen the class space for novel classes or pre-allocate a fixed class space. WaRP~\cite{kim2023iclr} transforms the old space into a new space to compress previous knowledge. ALICE~\cite{peng2022eccv} uses the angular penalty loss and NC-FSCIL~\cite{yang2023iclr} proposed a neural collapse-inspired classifier to pre-allocate the class space. However, it is difficult to accurately align the pre-allocated class space with new classes.

\begin{figure*}[t]
\centering
\includegraphics[width=0.95\linewidth]{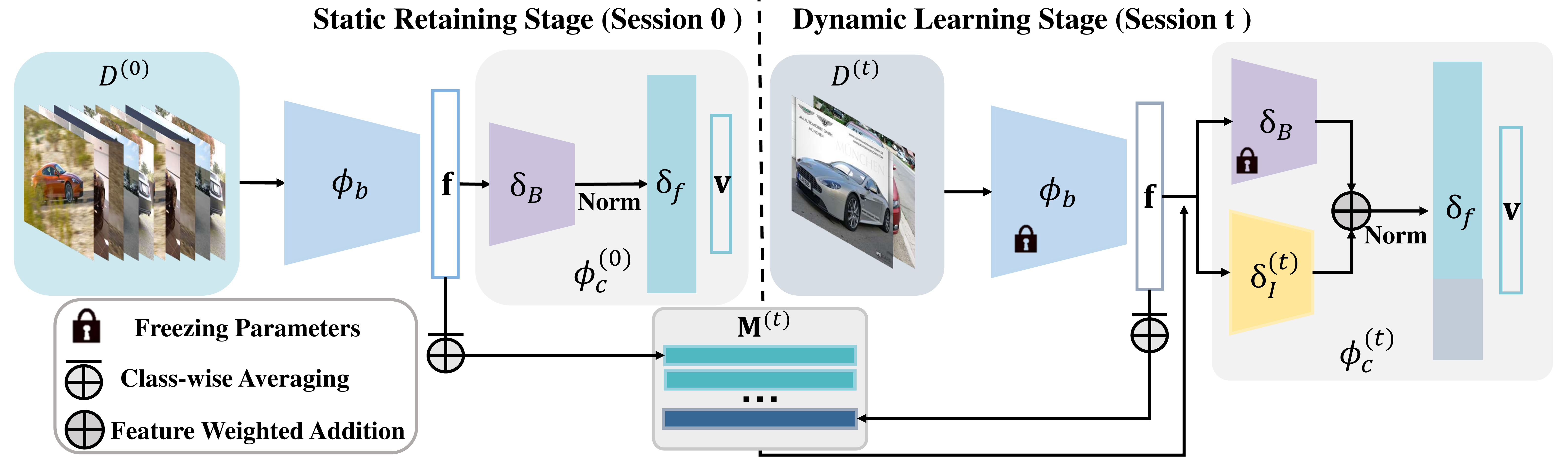}
\caption{Illustration of our framework (SDC). Left: The training process in the static retaining stage (session $0$). We train a model on the dataset $D^{(0)}$, which contains a backbone $\phi_b$ and a classifier $\phi_c^{(0)}$. The classifier consists of a projector $\delta_B$ and a fully connected layer $\delta_f$ for classification. Right: The training process in the dynamic learning stage (session $t (t>0)$). Before training, we freeze the backbone, retain the previous projector as a static projector $\delta_B$, and train a new dynamic projector $\delta_I^{(t)}$ with the assistance of a memory $\bl{M}^{(t)}$ and gain the classifier $\phi_c^{(t)}$. }
\label{fig:method}
\Description{Framework}
\end{figure*}

\myPara{Constrain the model in FSCIL.} This group of methods can be further divided into data-dependent methods and regular-dependent methods. Data-dependent methods save some samples or features in a memory. Some methods select real samples~\cite{rebuffi2017cvpr,lin2023tcsvt} from old classes replayed alongside new data, which are prone to exposing privacy. Some methods discourage forgetting and protect privacy by generating synthetic training examples~\cite{wu2018nips}, pseudo features~\cite{triki2017iccv,ostapenko2019cvpr} or prototypes~\cite{chen2021iclr,zhou2022cvpr}. These methods retain knowledge quickly, but there are still deviations between generated and real examples. Regular-dependent methods store architectures or constrain the parameter updating by a regularizer to promote memorization. LPILC~\cite{bai2022tcyb} uses a linear programming model to modify the classifier weights continuously. 
SPPR~\cite{zhu2021cvpr} adopts a randomly episodic training and the dynamic relation projection module. Some methods~\cite{zhao2023cvpr,cheraghian2021cvpr} introduce a knowledge distillation algorithm to draw knowledge from the teacher to the student. LDC~\cite{liu2023tpami} initializes biased distributions for all classes based on classifier vectors and a single covariance matrix. 
However, the success of these methods depends heavily on rationality designs, which relate to the performance in the current session.

\section{Preliminary}
nder dynamic environments, FSCIL~\cite{tao2020cvpr} trains a model upon a stream of training datasets $\mathcal{D}=\{D^{(t)}\}_{t=0}^{T}$. $D^{(t)}=\{(\bl{x}_i,y_i)\}_{i=1}^{|D^{(t)}|}$, where $\bl{x}_i$ is an example (e.g, image), $y_i\in L^{(t)}$ denotes its target. $L^{(t)}$ is its label space of session $t$. And for $\forall i, j$, $L^{(i)}\cap L^{(j)}=\varnothing$, which means there is no overlap in labeling space from different tasks. In FSCIL, $D^{(0)}$ is a training set in the base session containing many classes with sufficient training data for each class. $D^{(t)} (t>0)$ is a \emph{$N$-way $K$-shot} training set in the $t$-th incremental session, where has $N$ new classes and $K$ examples per class.  

Typically, we train a model $\phi(\bl{x})$ on $\mc{D}^{(0)}$ and further incrementally finetune it on $\mc{D}^{(t)} (t>0)$. And the model is evaluated on all previous validation datasets $\{V^{(i)}\}_{i=0}^t$. Our model $\phi(\bl{x})$ with the parameters $\theta$ is composed of two components: a backbone $\phi_b(\cdot)$ with parameters $\theta_b$, and a classifier $\phi_c(\cdot)$ with parameters $\theta_c$. Given an input image $\bl{x}_i$, the backbone extracts its intermediate feature as $\bl{f}_i$, and the classifier integrates the feature as $\bl{v}_i$. 
In the base session ($0$-th session), we train a model on dataset $D^{(0)}$, which optimize the parameters $\theta^{(0)} = \{\theta_b, \theta_c^{(0)}\}$. And in incremental sessions ($t>0$), we finetune parameters $\theta_c^{(t)}$ on dataset $D^{(t)} (t>0)$. 
Further, we use an extra memory $\bl{M}^{(t)}$ to promote memorization of old classes. The objective of FSCIL is minimizing the expected risk $\mathcal{R}$ into the base session $t=0$ (1a) and incremental sessions $t>0$ (1b):
\begin{subequations}\label{eq:problem}
\begin{align}
\min&\mathcal{R}_b(D^{(0)};\phi,\theta_b, \theta_c^{(0)}), \\
\min\mathcal{R}&_i(D^{(t)}, \bl{M}^{(t)};\phi,\theta_b, \theta_c^{(t)}).
\end{align}
\end{subequations}



\section{Methodology}

As shown in Figure~\ref{fig:method}, our framework is designed to effectively adapt new knowledge and retain previously acquired knowledge. The whole training encompasses two distinct yet interconnected training stages: static retaining stage (SRS) and dynamic learning stage (DLS). During SRS, we train a model on dataset $D^{(0)}$ in the base session. After training, we retain the projector from the base session as a static projector, which acts as a stable reference foundation for subsequent tasks. During DLS, we build a dynamic projector and train it on dataset $D^{(t)} (t>0)$ with the assistance of the static projector and an extra memory in incremental sessions, incrementally and plastically obtaining new information. 



\subsection{Static Retaining Stage}

In this phase, our primary objective is to refine the model using an abundant dataset, ensuring that it not only learns efficiently but also maintains a robust knowledge base. Based on this, we introduce a static retaining stage (\textbf{SRS}) to statically encapsulate fundamental knowledge and commonly recurring concepts, allowing it to retain this valuable information across subsequent sessions or iterations. This ensures that the model not only adapts to new data but also preserves its foundational understanding, facilitating continuous improvement while safeguarding essential knowledge.

\myPara{Model structure.} In the base session, the classifier $\phi_c^{(0)}$ consists of a projector $\delta_B$ and a fully connected (FC) layer $\delta_f^{(0)}$. The projector is an MLP block for projecting the features. And parameters of the FC layer $\delta_f^{(0)} $ is a $|L^{(0)}|$-dimensional matrix. Here, $\phi_c^{(0)} = \{\delta_B, \delta_f^{(0)}\}$. For a sample, its intermediate feature $\bl{f}_i$ from the backbone $\phi_b$ passes through the projector $\delta_B$, performs an $l_2$ normalization and is fed into the FC layer $\delta_f^{(0)}$ to obtain the output vector $\bl{v}_i$. The whole process is as follows 
\begin{equation}\label{eq:vector}
\bl{f}_i = \phi_b(\bl{x}_i), \bl{v}_i = \delta_f^{(0)}(\mathcal{N}(\delta_B(\bl{f}_i))),
\end{equation}
where $\mathcal{N}(\cdot)$ is the $l_2$ normalization.

\myPara{Training process.} When training, we feed samples into the backbone to generate intermediate features and then put them into the classifier to gain $L^{(0)}$-dimension vectors. The model are jointly trained on $D^{(0)}$ by minimizing the empirical risk loss as
\begin{equation}
    \mathcal{R}_{b} = \frac{1}{|D^{(0)}|}\sum_{(\bl{x}_i,y_i) \in D^{(0)}} \mathcal{L}_{ce}(\bl{v}_i,y_i),
\end{equation}
where $\mathcal{L}_{ce}$ is the cross-entropy loss. 

After training, we store the projector $\delta_B$ as a static projector for incremental learning. After adequately training on dataset $D^{(0)}$, the learned parameters and knowledge of the static projector ensure that the model learns the fundamental patterns and representations present, preserving the initial competency of the model and enabling it to maintain performance on base classes. The static projector maintains the fundamental knowledge from the base session, which serves as the cornerstone, providing a stable and reliable foundation upon which all subsequent learning can be built. It encapsulates some common knowledge and the core patterns of the base classes, which ensures that this essential knowledge remains intact and accessible throughout the learning process.




\subsection{Dynamic Learning Stage}

Mainstream methods use the strategy of freezing the backbone in incremental learning, which is superior to other incremental training operations based on fine-tuning strategies~\cite {yang2023iclr,yang2023tpami}. Therefore, this strategy is also utilized in our framework and we only finetune the classifier. The classifier undergoes continual refinement during incremental learning, which greatly improves the accuracy of the model. Consequently, we are dedicated to designing this crucial classifier and employ a dynamic learning stage (\textbf{DLS}). We introduce the static component to constrain the classifier's dynamical evolution, which enables the model to optimize the ability to adapt to new classes while mitigating catastrophic forgetting of previously learned classes, thereby better achieving the objectives of FSCIL.

\myPara{Model structure.} After SRS, the backbone is frozen, which is a common and effective technique under large model parameters and a few samples. We freeze the static projector and build a new dynamic projector similar to the static projector for projecting features with the same structure as the static projector. The FC layer is expanded to classify both old and new classes, where $\theta_f^{(t)}$ is a $\sum_{k=0}^{t} |L^{k}|$-dimensional matrix in $t$-th session. Here, $\phi_c^{(t)} = \{\delta_b, \delta_I^{(t)}, \delta_f^{(t)}\}$. In this situation, the vector $\bl{v}_i$ is processed by
\begin{equation}\label{eq:vector2}
\bl{v}_i = \delta_f^{(t)}\left(\mathcal{N}\left((1-\alpha)\delta_B(\bl{f}_i) + \alpha \delta_I^{(t)}(\bl{f}_i))\right)\right),
\end{equation}
where $\alpha$ is a factor to control the dominance of two projectors.

Besides, we also build a new FC layer to classify $\cup_{k=0}^{t} |L^{k}|$ classes, which can continue to grow with the session. Our model structure is thoughtfully designed to be both simple and lightweight, ensuring that it remains efficient while achieving high performance.

\myPara{Memory initialize and updating.} During incremental learning, we store the projector from the base session as the static projector, which is an important guarantee to retain knowledge of the base classes. However, for new classes from previous incremental sessions, the retention of knowledge is not guaranteed. Based on this, we employ an extra memory work together with our learning model during incremental sessions following~\cite{ostapenko2019cvpr,zhou2022cvpr,yang2023iclr}. Inspired by ProtoNet~\cite{snell2017nips}, the memory stores prototypes calculated by: 
\begin{equation}
    \label{eq:prototype}
    \begin{aligned}
    \bl{m}_{c}=\frac{1}{|D(y_i=l)|}\sum\nolimits_{\bl{x}_i \in D(y_i=l)}\phi_b(\bl{x}_i), l \in  L^{(t)}.
    \end{aligned}
\end{equation}

Before training, we initialize a memory $\bl{M}^{(1)}$ and calculate prototypes. All samples in $D^{(0)}$ are fed into the backbone, and then averaged by class into features by Eq.~(\ref{eq:prototype}), forming prototypes of the memory $\bl{M}^{(1)}$. And the length of it is $|L^{(0)}|$. After each incremental session, we update the memory $\bl{M}^{(t+1)}$. Similar to the initial, we calculate the prototypes by Eq. (\ref{eq:prototype}), which is acquired by averaging all features of the backbone for each class. Then we add them to the memory $\bl{M}^{(t+1)} (0< t < T-1)$. The length of the memory $\bl{M}^{(t+1)}$ is $\cup_{k=0}^{t} |L^{k}|$. In this way, each session $t$ uses the memory $\bf{M}^{(t)}$ for training the model, where $\bl{M}^{(t)} = \{(\bl{m}_{c},y_c)\}$ and $y_c \in \{L^{k}\}_{k=0}^{t}$.


\myPara{Training process.} In session $t (t>0)$, samples from dataset $D^{(t)}$ are passed into the frozen backbone to gain features. The features and the memory $\bf{M}^{(t)}$ are put into the classifier $\phi_c^{(t)}$, then we calculate vectors by Eq.~(\ref{eq:vector2}). After gaining vectors $\bl{v}_i$ and $\bl{v}_m$ from $\bl{f}_i$ and $\bl{m}_c$, we minimize the learning objective as 
\begin{equation}
\begin{split}
\mathcal{R}_{i} = &\frac{1}{|D^{(t)}|}\sum_{(\bl{x}_i,y_i) \in D^{(t)}} \mathcal{L}_{ce}(\bl{v}_i,y_i)\\
+&\frac{1}{|\bl{M}^{(t)}|}\sum_{(\bl{m}_c,y_c) \in \bl{M}^{(t)}} \mathcal{L}_{ce}(\bl{v}_c,y_c).
\end{split}
\label{eq:incloss}
\end{equation}

In DLS, we use the static projector and the dynamic projector for retaining old knowledge and learning new knowledge. The static projector is the foundation of incremental sessions, which projects unchanged samples and ensures the previous knowledge is powerfully preserved. The dynamic projector owns only a few parameters, which quickly adapt to new classes while reducing overfitting. In all, the cooperation between two projectors ensures that the framework has a reliable and consistent understanding of the starting point, and allows for seamless continuation and expansion of knowledge over time in harmony, fostering a balanced growth of knowledge. Besides, the double insurance of the memory and the static projectors powerfully preserved the old knowledge, ensuring the continuity of previous knowledge in subsequent sessions.

\begin{table*}[ht]
\centering
\caption{FSCIL performance on CIFAR100. ``Average accuracy'' means the average accuracy of all sessions and ``Average improvement'' calculates the improvement of our approach over other methods. 
These methods include class-incremental learning or few-shot learning methods with FSCIL setting and FSCIL methods. The best and second-best results are in bold and underlined, respectively.}
\resizebox{2.1\columnwidth}{!}{
\begin{tabular}{ccccccccccccc}
\hline
	\multirow{2}{*}{Method} & \multirow{2}{*}{Venue/Year} & \multicolumn{9}{c}{Accuracy in each session($\%$)$\uparrow$} & Average & Average \\
    & & 0 & 1 & 2 & 3 & 4 & 5 & 6 & 7 & 8 & accuracy & improvement\\
	\hline
 iCaRl \citep{rebuffi2017cvpr} & CVPR/2017 & 64.10 & 53.28 & 41.69 & 34.13 & 27.93 & 25.06 & 20.41 & 15.48 & 13.73 & 32.87 & +35.65  \\
EEIL \cite{castro2018eccv} & ECCV/2018 &	64.10&53.11 &43.71 	&35.15 &28.96& 	24.98 	&21.01 &	17.26 &	15.85 &	33.79 &	+34.77\\
\hline

TOPIC \citep{tao2020cvpr} & CVPR/2020 & 64.10 & 55.88 & 47.07 & 45.16 & 40.11 & 36.38 & 33.96 & 31.55 & 29.37 & 42.62 & +25.90  \\
LIMIT \citep{zhou2022tpami}& TPAMI/2022 & 73.81 & 72.09 & 67.87 & 63.89 & 60.70 & 57.77 & 55.67 & 53.52 & 51.23 & 61.84 & +6.68  \\
ALICE \citep{peng2022eccv} & ECCV/2022 & 79.00 & 70.50 & 67.10 & 63.40 & 61.20 & 59.20 & 58.10 & 56.30 & 54.10 & 63.21 & +5.31 \\
MCNet~\citep{ji2023tip} & TIP/2023 &73.30&69.34 &65.72&61.70&58.75 &56.44 &54.59 &53.01 &50.72 &60.40 &+7.78 \\
DSN \citep{yang2023tpami} & TPAMI/2023 & 73.00 & 68.83 & 64.82 & 62.64 & 59.36 & 56.96 & 54.04 & 51.57 & 50.00 & 60.14 & +8.04  \\
NC-FSCIL \citep{yang2023iclr}& ICLR/2023 & \underline{82.52} & \underline{76.82} & \underline{73.34} & \underline{69.68} & \underline{66.19} & \underline{62.85} & \underline{60.96} & \underline{59.02} & \underline{56.11} & \underline{67.50} & +1.02 \\
CABD~\citep{zhao2023cvpr} &  CVPR/2023 &79.45 	&75.38 	&71.84 	&67.95 	&64.96 	&61.95 	&60.16 	&57.67 	&55.88 	&66.14 	&+2.38 \\
OrCo~\citep{ahmed2024cvpr} & CVPR/2024	&80.08 	&68.16 	&66.99 	&60.97 	&59.78 	&58.60 	&57.04 	&55.13 	&52.19 	&62.10 	&+6.42 \\
OSHHG \citep{cui2024tmm} & TMM/2024 &63.55&62.88 &61.05 &58.13 &55.68 &54.59 &52.93 &50.39 &49.48 &56.52 &+12.00 \\
EHS \citep{deng2024wacv} & WACV/2024 &71.27&67.40 &63.87 &60.40 &57.84 &55.09 &53.10 &51.45 &49.43 &58.87 &+9.65 \\
         \hline    
\textbf{SDC (Ours)} & - & \textbf{83.40} & \textbf{79.62} & \textbf{74.31} & \textbf{70.31} & \textbf{67.70} & \textbf{63.45} & \textbf{61.36} & \textbf{59.45} & \textbf{57.05} & \textbf{68.52} & - \\
\hline
\end{tabular}
}
\label{table1}
\end{table*}

\subsection{Analysis} 

Our method is designed to decompose the core objective into two complementary components, enabling more efficient achievement of our goals. To further elucidate the integration of static and dynamic projectors, we analyze the optimization of Eq.~(\ref{eq:incloss}) from an information-theoretic perspective.
Mutual information measures the dependence between two random variables. For random variables $X$ and $Y$, the mutual information is $I(X,Y) = H(X) - H(X|Y)$, 
where $H(X)$ is the entropy of $X$, and $H(X|Y)$ is the conditional entropy of $X$ given $Y$. A higher mutual information indicates a stronger dependence between $X$ and $Y$.

For the vector $s_{i}=\delta_B(\bl{f}_i)$ from the static projector and the vector  $d_{i}=\delta_I(\bl{f}_i)$ from the dynamic projector, the mutual information is :
\begin{equation}
    \label{eq:SDCmi}
    \begin{aligned}
    I(s_i,v_i) = H(s_i) - H(s_i|v_i), I(d_i,v_i) = H(d_i) - H(d_i|v_i).
    \end{aligned}
\end{equation}
The static projector ensures that the model retains a stable representation of old knowledge, preventing catastrophic forgetting, while the dynamic projector enables the model to efficiently adapt to new tasks, enhancing its plasticity. Combining these with Eq.~(\ref{eq:vector2}), the overall optimization objective is:
\begin{equation}
    \label{eq:SDCop}
    \begin{aligned}
    \arg\max \left((1-\alpha)I(s_{i}|v_i) + \alpha I(d_{i}|v_i)\right).
    \end{aligned}
\end{equation}
The $\alpha$ controls the trade-off between stability (retaining old knowledge) and plasticity (adapting to new knowledge). By optimizing this objective, the model achieves a harmonious balance between these two competing goals, enabling it to simultaneously retain old knowledge and adapt to new tasks. This effectively addresses the stability-plasticity dilemma in few-shot class-incremental learning.


\section{Experiments}

\subsection{Experimental Settings}
\myPara{Datasets and split.} 
Following the common benchmark setting~\citep{tao2020cvpr,zhang2021cvpr,yang2023iclr}, we conduct experiments to evaluate the proposed SDC on three commonly used datasets, including CIFAR100~\cite{2009Learning}, MiniImageNet~\cite{russakovsky2015ijcv} and CUB200~\cite{Wah2011TheCB}. 
MiniImageNet and CIFAR100 consist of 100 classes, where 60 classes are set for the base session and 40 classes are set for incremental sessions. Each incremental session is a \emph{$5$-way $5$-shot} setting, which means each session has 5 new classes and each new class has 5 samples. CUB200 consists of 200 classes, where 100 classes are used in the base session and the other 100 classes are divided into 10 different sets with a \emph{$10$-way $5$-shot} setting in incremental sessions. And we also conduct experiments on an application dataset, FGVC-Aircraft~\cite{DBLP:journals/corr/MajiRKBV13}. FGVC-Aircraft has 10,200 images covering 102 classes of aircraft variants. Following~\cite{goswami2024cvpr}, we randomly select 100 classes for training. Similar to CIFAR100 and MiniImageNet, we select 60 classes for the base session and 40 classes for incremental sessions with a \emph{$5$-way $5$-shot} setting. 

\begin{table*}[ht]
\centering
\caption{FSCIL performance on MiniImageNet. ``Average accuracy'' means the average accuracy of all sessions and ``Average improvement'' calculates the improvement of our approach over other methods. These methods include class-incremental learning or few-shot learning methods with FSCIL setting and FSCIL methods. The best and second-best results are in bold and underlined, respectively.}
\resizebox{2.1\columnwidth}{!}{
\begin{tabular}{ccccccccccccc}
\hline
	\multirow{2}{*}{Method} & \multirow{2}{*}{Venue/Year} & \multicolumn{9}{c}{Accuracy in each session($\%$)$\uparrow$} & Average & Average \\
    & & 0 & 1 & 2 & 3 & 4 & 5 & 6 & 7 & 8 & accuracy & improvement\\
	\hline
iCaRl \citep{rebuffi2017cvpr} & CVPR/2017 & 61.31 & 46.32 & 42.94 & 37.63 & 30.49 & 24.00 & 20.89 & 18.80 & 17.21 & 33.29 & +35.45  \\
EEIL \cite{castro2018eccv} & ECCV/2018 & 61.31 & 46.58 & 44 & 37.29 & 33.14 & 27.12 & 24.1 & 21.57 & 19.58 & 34.97  & +33.77 \\
        \hline
        TOPIC \citep{tao2020cvpr} & CVPR/2020 & 61.31 & 50.09 & 45.17 & 41.16 & 37.48 & 35.52 & 32.19 & 29.46 & 24.42 & 39.64 & +29.10  \\
        LIMIT \citep{zhou2022tpami} &TPAMI/2022 & 72.32 & 68.47 & 64.30 & 60.78 & 57.95 & 55.07 & 52.70 & 50.72 & 49.19 & 59.06 & +9.68  \\
        C-FSCIL \citep{hersche2022cvpr} & CVPR/2022 & 76.40 & 71.14 & 66.46 & 63.29 & 60.42 & 57.46 & 54.78 & 53.11 & 51.41 & 61.61 & +7.13 \\
        ALICE \citep{peng2022eccv} & ECCV/2022& 80.60 & 70.60 & 67.40 & 64.50 & 62.50 & 60.00 & 57.80 & 56.80 & 55.70 & 63.99 & +4.75 \\
        MCNet~\citep{ji2023tip} & TIP/2023 &72.33&67.70 &63.50 &60.34 &57.59 &54.70 &52.13 &50.41 &49.08 &58.64 &+10.10 \\
        NC-FSCIL \citep{yang2023iclr} &ICLR/2023 & \underline{84.02} & \underline{76.80} & \underline{72.00} & \underline{67.83} & \underline{66.35} & \underline{64.04} & \underline{61.46} & \underline{59.54} & \underline{58.31} & \underline{67.82} & +0.92 \\
        CABD~\citep{zhao2023cvpr}&CVPR/2023	&74.65 	&70.43 	&66.29 	&62.77 	&60.75 	&57.24 	&54.79 	&53.65 	&52.22 	&61.42 	&+7.32 \\
        OrCo~\citep{ahmed2024cvpr} & CVPR/2024	&83.30 	&75.32 	&71.53 	&68.16 &	65.63 	&63.12 	&60.20 	&58.82 	&58.08 	&67.13 	&+1.61 \\
        OSHHG \citep{cui2024tmm} & TMM/2024 &60.65&59.00 &56.59 &54.78 &53.02 &50.73 &48.46 &47.34 &46.75 &53.04 &+15.70\\
        EHS \citep{deng2024wacv} & WACV/2024 &69.43&64.86 &61.30 &58.21 	&55.49 	&52.77&50.22 &48.61 &47.67 &56.51 &+12.23\\
        \hline
         \textbf{SDC (Ours)} & - & \textbf{84.89} &\textbf{78.42} &\textbf{73.57} &\textbf{68.81} &\textbf{67.07} &\textbf{64.74} &\textbf{62.25} &\textbf{60.13} &\textbf{58.75}  & \textbf{68.74} & - \\
         \hline
\end{tabular}
}
\label{table2}
\end{table*}

\begin{table*}[ht]
\centering
\caption{FSCIL performance on CUB200. ``Average accuracy'' means the average accuracy of all sessions and ``Average improvement'' calculates the improvement of our approach over other methods. These methods include class-incremental learning or few-shot learning methods with FSCIL setting and FSCIL methods. The best and second-best results are in bold and underlined, respectively.}
\resizebox{2.1\columnwidth}{!}{
\begin{tabular}{ccccccccccccccc}
\hline
	\multirow{2}{*}{Method}& \multirow{2}{*}{Venue/Year} & \multicolumn{11}{c}{Accuracy in each session($\%$)$\uparrow$} & Average & Average \\
    & & 0 & 1 & 2 & 3 & 4 & 5 & 6 & 7 & 8 & 9 & 10 & accuracy & improvement\\
	\hline
        iCaRl \citep{rebuffi2017cvpr}& CVPR/2017 & 68.68 & 52.65 & 48.61 & 44.16 & 36.62 & 29.52 & 27.83 & 26.26 & 24.01 & 23.89 & 21.16 & 36.67 & +30.93   \\
        EEIL\cite{castro2018eccv} & ECCV/2018 & 68.68  & 53.63  & 47.91  & 44.20  & 36.30  & 27.46  & 25.93  & 24.70  & 23.95  & 24.13  & 22.11  & 36.27  & +31.33 \\
       \hline
        TOPIC \citep{tao2020cvpr} & CVPR/2020 & 68.68 & 62.49 & 54.81 & 49.99 & 45.25 & 41.40 & 38.35 & 35.36 & 32.22 & 28.31 & 26.28 & 43.92 & +23.68 \\
        LIMIT\citep{zhou2022tpami} & TPAMI/2022 & 76.32 & 74.18 & 72.68 & 69.19 & \textbf{68.79} & 65.64 & 63.57 & 62.69 & \underline{61.47} & 60.44 & 58.45 & 66.67 & +0.93\\
        ALICE \citep{peng2022eccv}&ECCV/2022 & 77.40 & 72.70 & 70.60 & 67.20 & 65.90 & 63.40 & 62.90 & 61.90 & 60.50 & \underline{60.60} & \underline{60.10} & 65.75 & +1.85\\
        MCNet~\citep{ji2023tip}&TIP/2023 &77.57 &73.96 &70.47 &65.81 &66.16 &63.81 &62.09 &61.82 &60.41 &60.09 &59.08 &65.57  &+2.03 \\
      NC-FSCIL \citep{yang2023iclr}&ICLR/2023 & \underline{80.45} & \underline{75.98} & {72.30} & \underline{70.28} & {68.17} & \underline{65.16} & \textbf{64.43} & 63.25 & {60.66} & {60.01} & {59.44} & {67.28} & +0.32\\
      CABD~\citep{zhao2023cvpr}&CVPR/2023&79.12 	&75.37 &	\underline{72.80} 	&69.05 	&67.53 	&65.12 	&64.00 	&\textbf{63.51} 	&\textbf{61.87} 	&\textbf{61.47} 	&\textbf{60.93}	&\textbf{67.34} 	&+0.26 \\
    OrCo~\citep{zhao2023cvpr}	&CVPR/2024&75.59 	&66.85 	&64.05 	&63.69 	&62.20 	&60.38 	&60.18 	&59.20 	&58.00 	&58.42 	&57.94 &	62.41 	&+5.19 \\
      MgSvF \citep{zhao2024tpami} &TPAMI/2024& 72.29 & 70.53 & 67.00 & 64.92 & 62.67 & 61.89 & 59.63 & 59.15 & 57.73 & 55.92 & 54.33 & 62.37 & +5.23 \\
      OSHHG \citep{cui2024tmm} &TMM/2024&63.20 &62.61 &59.83 &56.82 &55.07 &53.06 &51.56&50.05&47.50&46.82&45.87&53.85 &+13.75\\
      \hline
\textbf{SDC (Ours)}& - & \textbf{80.78} &\textbf{76.72} &\textbf{73.18} &\textbf{70.98} &\underline{68.52} &\textbf{65.73}& \underline{64.26}& \underline{63.31}& 60.84& 60.07 & 59.24 & \textbf{67.60} & -\\
\hline
\end{tabular}
}
\label{table3}
\end{table*}

\begin{table*}[ht]
\centering
\caption{FSCIL performance on FGVC-Aircraft. ``Average accuracy'' means the average accuracy of all sessions and ``Average improvement'' calculates the improvement of our approach over other approaches. These approaches include class-incremental learning with FSCIL setting and FSCIL approaches. The best results are in bold. On this dataset, we reproduce some representative SOTA methods to compare with our SDC.}
\resizebox{2.0\columnwidth}{!}{
\begin{tabular}{cccccccccccc} 
\hline
	\multirow{2}{*}{Method} & \multicolumn{9}{c}{Accuracy in each session($\%$)$\uparrow$} & Average & Average \\
    &  0 & 1 & 2 & 3 & 4 & 5 & 6 & 7 & 8 & accuracy & improvement\\
	\hline
        EEIL \cite{castro2018eccv} & 70.32 & 55.31 & 49.28 & 45.87 & 38.11 &	28.56  &26.83  &25.77  &25.45  & 40.61 & +26.74  \\
        C-FSCIL \cite{hersche2022cvpr} & 82.55 & 74.33 	& 71.88 	& 68.72 	& 65.34 & 62.16 & 59.19 & 56.14 & 52.76 & 65.79 & +1.56  \\ 
        NC-FSCIL~\cite{yang2023iclr} & 81.70 & 74.97  & 72.14  & 69.23  & 65.72  & 63.01  & 59.93  &57.67  &53.23  &66.40  &	+0.95 \\
	 \textbf{ SDC (Ours)} & \textbf{82.55} 	&\textbf{75.66} 	&\textbf{73.00} 	&\textbf{70.12} &\textbf{	66.39} 	&\textbf{64.73} 	&\textbf{61.33} 	&\textbf{58.00} & \textbf{54.37} & \textbf{67.35} &- \\
\hline
\end{tabular}
}
\label{table4}
\end{table*}

\myPara{Networks.} For fair comparisons, we follow the same setting in other FSCIL methods \cite{zhang2021cvpr,yang2023iclr} and use Resnet~\cite{he2016cvpr} as the backbone. Specially, we use Resnet12 without pretraining for CIFAR100, MiniImageNet, and FGVC-Aircraft, and Resnet18 pre-trained on ImageNet for CUB200. Additionally, we use a two-layer MLP as the projector for feature transformation.

\myPara{Implementation.} To train models, we use SGD with a momentum of $0.9$ as the optimizer throughout all sessions. We adopt Mixup~\cite{zhang2018iclr} and CutMix~\cite{yun2019iccv} as augmentation schemes and perform random translation, rotation, and contrast enhancement. The input images are rescaled into $32\times32$ for CIFAR100, $84\times84$ for MiniImageNet and $224\times224$ for CUB200 and FGVC-Aircraft,  which is the same as other methods. The batch size is set to 512 during training. On MiniImageNet and CIFAR100, we train for 100 epochs in the base session, and $100-300$ epochs in each incremental session, with an initial learning rate of $0.25$ for all sessions. On CUB200, we train for $200$ epochs in the base session, and $200-300$ epochs in each incremental session, with an initial learning rate of $0.01$ for the base session and $0.001$ for incremental sessions. On FGVC-Aircraft, we train for $100$ epochs in the base session, and $300$ epochs in each incremental session, with an initial learning rate of $0.25$ for the base session and $0.025$ for incremental sessions. 

\subsection{State-of-the-art Comparison}

We conduct the experimental comparisons with the state-of-the-art methods and report the evaluated accuracy of each session on CIFAR100, MiniImageNet, CUB200, and FGVC-Aircraft as shown in Table~\ref{table1}, Table~\ref{table2}, Table~\ref{table3} and Table~\ref{table4}. We analyze all results in two parts: (1) analyze the trends of accuracy with the increase of sessions, and (2) compare our SDC with some SOTA methods.

With the model processing the continuous stream of new tasks, the accuracy of all sessions shows a clear downward trend. On the one hand, the model needs to learn new knowledge based on previous sessions and a small number of samples, which means the model needs to identify more samples and classes. On the other hand, the degradation is triggered by catastrophic forgetting. Specifically, when the model learns a new task, its internal parameters and structure adjust accordingly to accommodate the new data distribution and patterns. Adjustments always affect the ability to process new knowledge and align it with old knowledge. 

Observing the results in Table~\ref{table1}, Table~\ref{table2}, Table~\ref{table3} and Table~\ref{table4}, our framework has performance advantages compared with related methods, which effectively recognizes novel classes incrementally and effectively mitigates catastrophic forgetting.
Specifically, our framework surpasses some class-incremental learning (CIL) methods (such as iCaRL~\cite{rebuffi2017cvpr}, EEIL~\cite{castro2018eccv}), which handle catastrophic forgetting in Class-Incremental Learning (CIL) rather than FSCIL. The methods above are susceptible to the absence of instances, which exist in the overfitting phenomenon. And compared to current FSCIL methods, our framework achieves excellent performance on all datasets. Our framework stands out among many alternatives and achieves the best performance in all sessions on CIFAR100, MiniImageNet and FGVC-Aircraft, which achieve the average accuracy of $68.60\%$ on CIFAR100, $68.74\%$ on MiniImageNet, and $67.35\%$ on FGVC-Aircraft. With the optimization of the model and data processing, we extract valuable information, which improves accuracy and efficiency. Our framework can also continuously learn new knowledge while retaining old knowledge in real-world applications.
As we obtain the average accuracy of $67.60\%$ on CUB200. Although the accuracy curve was dented on some sessions on CUB200, we still achieved the highest average accuracy. 

\subsection{Ablation Study}

\begin{table*}[ht]
\centering
\caption{The effect of $\alpha$ on CIFAR100, which control the dominance of two projectors. The best results are in bold.}
\resizebox{1.3\columnwidth}{!}{
\begin{tabular}{cccccccccc} 
\hline
	\multirow{2}{*}{$\alpha$} & \multicolumn{9}{c}{Accuracy in each session($\%$)$\uparrow$} \\
    &  0 & 1 & 2 & 3 & 4 & 5 & 6 & 7 & 8\\
	\hline
        0 & \multirow{7}{*}{\textbf{83.40}} & 78.17 & 72.56 & 67.16 & 62.88 & 57.93 &	55.86  &53.13  & 50.95 \\
        $0.1$ & & 76.12 & 71.97 & 68.62 & 65.24 & 62.26 & 58.87 & 55.24 & 52.36  \\ 
        $0.3$ & & 79.11 & 73.54 & 69.17 & 66.31 &	62.58  &60.43  &57.67  &54.67 \\
        \textbf{0.5} & & \textbf{79.62} & \textbf{74.31} & \textbf{70.31} & \textbf{67.70} & \textbf{63.45} & \textbf{61.36} & \textbf{59.45} & \textbf{57.05}  \\ 
        $0.7$ &  & 79.08 & 73.83 & 69.78 & 66.13 &62.92  &60.78  &58.21  &54.78 \\
        $0.9$ &  & 78.68 & 72.36 	& 69.02 	& 65.75 & 62.63 & 59.29 & 56.14 & 54.51  \\ 
        $1$ & & 78.36 & 71.78 & 68.97 & 65.11 &	62.43  &58.93  &55.84  &54.15 \\        
\hline
\end{tabular}
}
\label{table5}
\end{table*}


\begin{table}[t]
\centering
\caption{The effect of layer number and different dimensions on CIFAR100. ``Dimension'' denotes using different output dimensions of each projector. ``Base'', ``First'', and ``Final'' are the accuracy of the base session, the first incremental session, and the last incremental session, respectively. The best results in all lines are in bold.}
\resizebox{0.95\columnwidth}{!}{
\begin{tabular}{c|c|cccc}
\hline
\multirow{2}{*}{Dimension} & \multirow{2}{*}{Session} & \multicolumn{4}{c}{Layer number} \\ 
&&1 & 2 & 3 & 4\\
\hline
\multirow{3}{*}{512} & Base &83.47& 83.67 & \textbf{84.00}&	83.57\\
&First&78.83 &79.01 &\textbf{79.53} &78.82 \\
&Final&55.89 &\textbf{55.87} & 55.42& 55.00\\
\hline
\multirow{3}{*}{1024} & Base & 84.01 &84.07 & 84.12&\textbf{84.57}\\
&First&79.09 &\textbf{79.31} &79.04 &78.95 \\
&Final&55.21 &\textbf{56.37} & 56.07 &55.60\\
\hline
\multirow{3}{*}{2048} & Base &83.77 &84.40 &\textbf{84.55} &84.03\\
&First&78.89 &\textbf{79.62} &79.23 & 78.98\\
&Final& 55.68& \textbf{56.95} & 56.39&55.23 \\
\hline
\end{tabular}
}
\label{ab}
\end{table}
\begin{figure*}[ht]
\centering
\includegraphics[width=1.0\linewidth]{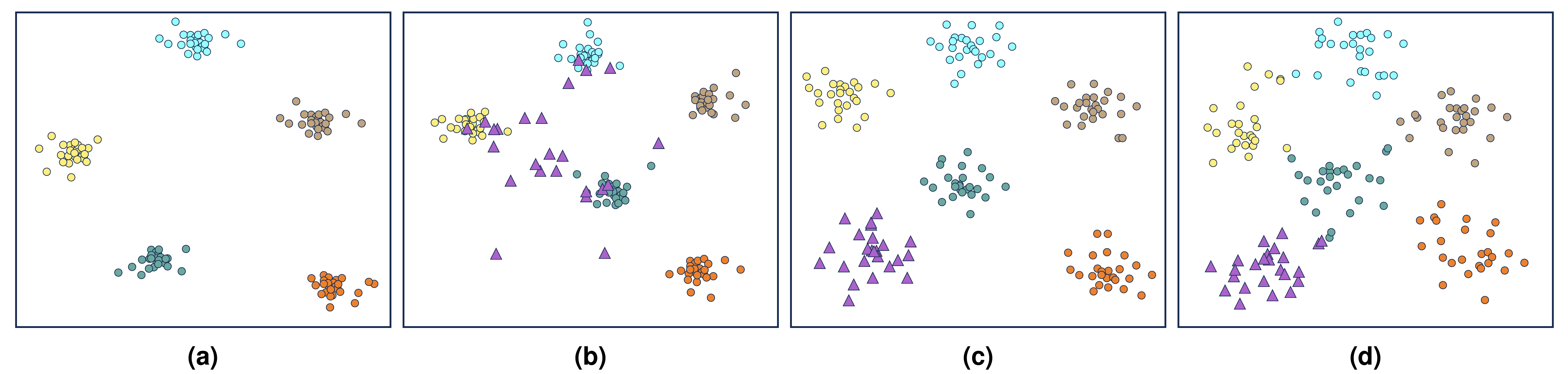}
\caption{Representation visualization with t-SNE~\cite{2008Visualizing}, which uses session $0$ and session $1$ on CIFAR100 as an example. We randomly select 25 examples over 5 base classes and 1 incremental class to show the model effect. Symbols `$\bullet$' and `$\blacktriangle$' represent examples of base classes and incremental classes, respectively. (a) visual features of base classes from session $0$. (b) , (c), and (d) are visual features of all classes when $\alpha = 0$, $\alpha = 0.5$, and $\alpha = 1$ from session $1$, respectively.}
\label{fig:tsne}
\Description{tsne}
\end{figure*}

\myPara{Effect of $\alpha$.} As shown in Table~\ref{table5}, our SDC achieves the best accuracy when $\alpha = 0.5$, meaning static and dynamic projectors are equally important in the training process. Among them, the static projector provides a more reliable foundation for models by storing knowledge statically, even as new data and tasks are introduced. The dynamic projector assimilates new patterns and knowledge actively and continuously, evolving its adaptability to new data. While the static projector ensures stability and reliability, the dynamic projector fosters adaptability and innovation. By assigning equal weights to these two projectors, the model can harness complementary strengths of the two projectors effectively. 
When we only use the static projector ($\alpha = 0$), the model has high accuracy at the beginning of incremental sessions due to the strong ability to identify base classes. When we only use the dynamic projector ($\alpha = 1$), the model has a better accuracy at the end. When the static projector assumes a more prominent role within the model ($0<\alpha<0.5$), the dynamic projector still impacts the accuracy. When the dynamic projector dominates ($0.5<\alpha<1$), the model performs better at the end due to being more sensitive to new classes.

\myPara{Effect of layer number.} Table~\ref{ab} compares the accuracy of different layer numbers per projector. In the base session, the model achieves the best accuracy when using a three-layer MLP as the projector under $512$ dimensions and $2048$ dimensions, and using a four-layer MLP as the projector under $1024$ dimensions. The performance correlates positively with the layer number of the projector. This phenomenon is due to the projector with a greater number of parameters can encapsulate and retain richer knowledge when having sufficient data. While increasing the layer number enhances performance up to a certain point, it eventually reaches a saturation point due to the model overfitting to the training data. Therefore, further expanding the model leads to decreased performance on test sets. In incremental learning, the model obtains the best accuracy when using a two-layer MLP as the projector under $1024$ dimensions and $2048$ dimensions. For $512$ dimensions, it also gains the best accuracy for previous sessions when using a three-layer MLP, and obtains the best accuracy for subsequent sessions when using a two-layer MLP. Indeed, the model faces a significant challenge when adjusting a large number of parameters with a few samples at this stage. The primary reason lies that limited samples hinder the effective optimization of a complex model. Considering the balance between model performance and resource consumption, we use a two-layer MLP as the projector during the whole training process. Compared to other methods, we use fewer parameter layers.

\myPara{Effect of output dimension.} As shown in Table~\ref{ab}, the dimension and the layer number of the projector are closely related to the performance. Within a certain number of parameters, the higher the model dimension and the better the model performance. As these two factors increase, so does the number of parameters that the model needs to learn, enabling it to capture more intricate patterns and relationships within the data. When dealing with only a limited number of samples, the abundance of parameters always causes the model to overfit and limits the ability of the model to adapt and learn new knowledge effectively. Considering the number of layers and dimensions of the projector, we use $2048$ for the training. 

\subsection{Further Discussion}


Moreover, to further demonstrate that the cooperation of static and dynamic projectors can better overcome catastrophic forgetting and learn incremental classes, we introduce the widely used t-SNE~\cite{2008Visualizing} tool to visualize feature distribution. We visualize the embedding space on the CIFAR100 in Figure~\ref{fig:tsne}. As can be observed from the figure, our approach allows for more precise separation of different classes and tighter clustering of the same class. 
Firstly, the samples of the same class can be aggregated well except for some special cases, where the model can effectively capture common patterns of the same class and distinguish the intrinsic properties of the different classes. Secondly, novel classes have closer inter-class distances than base classes, which means learning new data is more difficult than learning base data, resulting in the model being less sensitive to the differences between novel classes. Finally, synthetic samples are well aggregated in their classes, where the model can learn from real data and effectively transfer this learning to the synthetic set. Meanwhile, this also indicates that the model successfully captures the real data's key characteristics and saves them in simulated samples for subsequent tasks. 
When we only use the static projector ($\alpha = 0$), the model imposes stronger constraints on old classes through fixed projection parameters, forcing their embeddings to maintain compact aggregation within the original feature space. When we only use the dynamic projector ($\alpha = 1$), it enhances new knowledge absorption by adaptively adjusting projection parameters during incremental learning, enabling better separation of emerging classes.


\section{Conclusion}

In this paper, we propose a framework termed Static-Dynamic Collaboration (\textbf{SDC}) to facilitate FSCIL. During SRS, we train a model and store the projector as a static projector in the base session. During DLS, we freeze the backbone and train a new dynamic projector working with the static projector and an extra memory in incremental sessions. The interplay between two projectors effectively balances the preservation of old knowledge with the acquisition of new insights, achieving effective model learning. Extensive experiments and ablation studies clearly show the effectiveness, universality and simplicity of our SDC. Our SDC significantly enhances the capabilities of the model, enabling it to learn more efficiently and adapt to new situations more effectively, which can be used more widely and generally in many practical scenarios.  

\begin{acks}
This work was partially supported by grants from the Pioneer R$\&$D Program of Zhejiang Province (2024C01024), and Open Research Project of National Key Laboratory
of Science and Technology on Space-Born Intelligent Information Processing (TJ-02-
22-01).
\end{acks}


\bibliographystyle{ACM-Reference-Format}
\balance
\bibliography{icmr}

\end{document}